\PassOptionsToPackage{hyphens}{url} % temp fix for imcompatible url imports issue
\documentclass[10pt,twocolumn,letterpaper]{article}

\usepackage{color}
\usepackage{epsfig}
\usepackage{graphicx}

\usepackage{adjustbox}
\usepackage{array}
\usepackage{booktabs}
\usepackage{colortbl}
\usepackage{wrapfig}
\usepackage{hhline}
\usepackage{multirow}
\usepackage{wrapfig}
\usepackage{floatflt}

\usepackage{amsmath,amsfonts,amssymb}
\usepackage{mathtools}  % additional useful math commands, such as \coloneqq
\usepackage{bm}
\usepackage{nicefrac}
\usepackage{microtype}
\usepackage{dsfont}
\usepackage{bbm}
\usepackage{markdown}

\usepackage{changepage}
\usepackage{extramarks}
\usepackage{fancyhdr}
\usepackage{lastpage}
\usepackage{setspace}
\usepackage{soul}
\usepackage{xspace}

\usepackage{makecell}

\usepackage{pifont} % for extra symbols such as \cmark

\usepackage{algorithm,algpseudocode}

\usepackage{amsthm}
\usepackage[symbol]{footmisc}
\renewcommand{\thefootnote}{\fnsymbol{footnote}}

\newcolumntype{L}[1]{>{\raggedright\let\newline\\\arraybackslash\hspace{0pt}}m{#1}}
\newcolumntype{C}[1]{>{\centering\let\newline\\\arraybackslash\hspace{0pt}}m{#1}}
\newcolumntype{R}[1]{>{\raggedleft\let\newline\\\arraybackslash\hspace{0pt}}m{#1}}

\newcommand{\ignore}[1]{}

\DeclareMathOperator*{\argmin}{arg\,min}

\makeatletter
\DeclareRobustCommand\onedot{\futurelet\@let@token\@onedot}
\def\@onedot{\ifx\@let@token.\else.\null\fi\xspace}

\makeatother

\definecolor{MyDarkBlue}{rgb}{0,0.08,0.8}
\definecolor{MyDarkGreen}{RGB}{45,155,45}
\definecolor{MyDarkRed}{rgb}{0.8,0.02,0.02}
\definecolor{MyOrange}{rgb}{1.0, 0.4, 0.2}
\definecolor{MyPurple}{RGB}{111,0,255}
\definecolor{MyRed}{rgb}{0.8,0.0,0.0}
\definecolor{MyGold}{rgb}{0.75,0.6,0.12}
\definecolor{MyDarkgray}{rgb}{0.66, 0.66, 0.66}

\newcommand{\myparagraph}[1]{\vspace{5pt}\noindent\textbf{#1.}}

\usepackage[pagenumbers]{cvpr}
\usepackage{tabularx}
\definecolor{cvprblue}{rgb}{0.21,0.49,0.74}
\usepackage[pagebackref,breaklinks,colorlinks,allcolors=cvprblue]{hyperref}
\hypersetup{
    colorlinks=true,
    linkcolor=blue!70!black,
    citecolor=blue!70!black,
    filecolor=blue!70!black,
    urlcolor=blue!70!black  % This changes the color of \href and \url
}
\def\paperID{5846} % *** Enter the Paper ID here
\def\confName{CVPR}
\def\confYear{2025}

\newcommand{\method}{\textsc{LayoutVLM}\xspace}
\definecolor{lightergray}{gray}{0.9} % Define a lighter gray (adjust 0.95 for desired lightness)
\usepackage{listings} % For syntax highlighting
\usepackage{mdframed} % For styling frames

\usepackage[T1]{fontenc} % Improves underscore rendering
\usepackage{textcomp}     % Handles special characters better

\newcommand{\styledfileinput}[3][text]{% #1: language, #2: file, #3: additional options
  \begin{mdframed}[backgroundcolor=gray!10,linewidth=0pt]
    \lstinputlisting[language=#1,#3]{#2} % Syntax-highlighted content with extra options
  \end{mdframed}
}

\title{\method: Differentiable Optimization of 3D Layout \\via Vision-Language Models}
\author{Fan-Yun Sun*$^{1}$, Weiyu Liu*$^{1}$, Siyi Gu$^{1}$, Dylan Lim$^{1}$, \\
Goutam Bhat$^{2}$, Federico Tombari$^{2}$, Manling Li$^{1}$, Nick Haber$^1$, Jiajun Wu$^{1}$ \vspace{0.1cm} \\\\
$^1$Stanford University, $^2$Google Research \vspace{0.1cm} \\
{\small \textcolor{blue!70!black}{\tt \href{https://ai.stanford.edu/~sunfanyun/layoutvlm/}{https://ai.stanford.edu/\textasciitilde sunfanyun/layoutvlm/}}} \vspace{-0.2cm}
}

\begin{document}
\twocolumn[{%
\maketitle

%\vspace{-1cm}
\begin{figure}[H]
\hsize=\textwidth
    \centering
   \includegraphics[width=\textwidth]{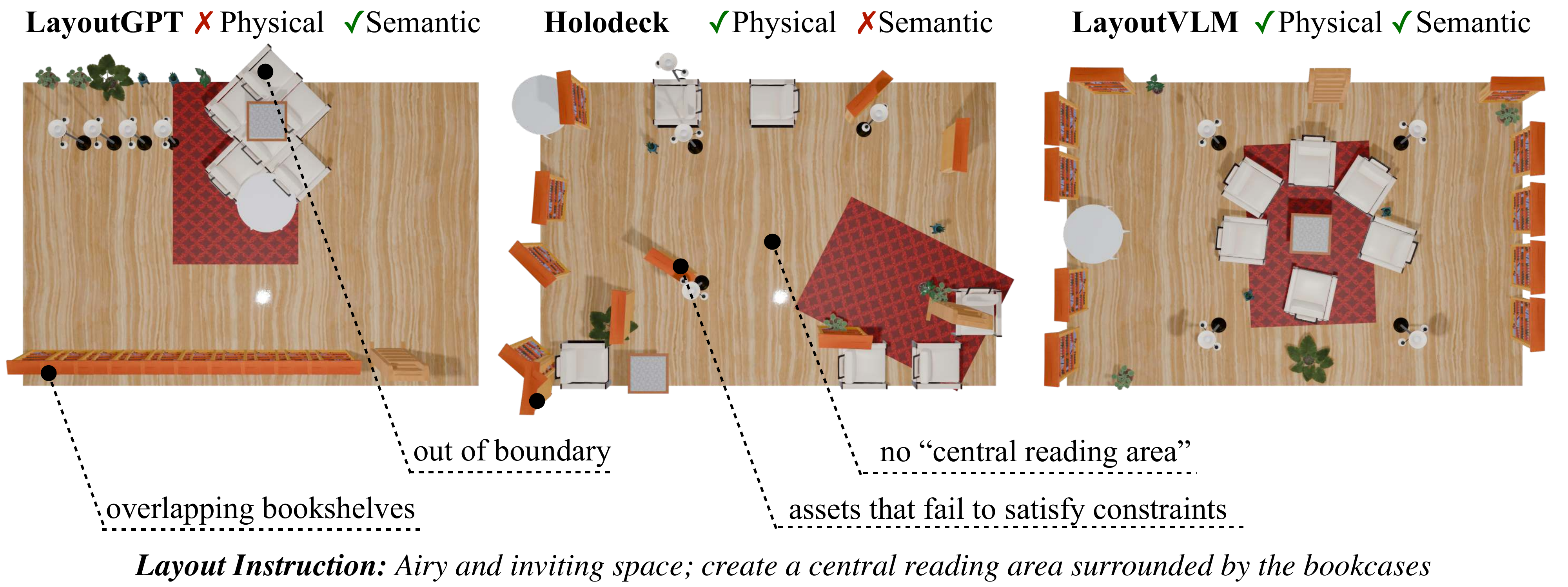}
   %\vspace{-0.5cm}
    \caption{From \textbf{unlabeled} 3D assets and language instruction, \method generates scene layouts that are physically plausible and semantically coherent—two criteria that existing methods often struggle to meet. Our approach addresses this by using a VLM to generate a scene layout representation that defines both an initial layout and spatial relations between assets for differentiable optimization.}
    \label{fig:pull}
\end{figure}
}]

\renewcommand{\thefootnote}{\fnsymbol{footnote}}\footnotetext[1]{Equal contribution.}\renewcommand{\thefootnote}{\arabic{footnote}}\setcounter{footnote}{0}

\begin{abstract}
%Open-universe 3D layout generation arranges unlabeled 3D assets conditioned on language instruction. Large language models (LLMs) struggle with generating physically plausible 3D scenes and adherence to input instructions, particularly in dense scenes. 
Spatial reasoning is a fundamental aspect of human cognition, enabling intuitive understanding and manipulation of objects in three-dimensional space. While foundation models demonstrate remarkable performance on some benchmarks, they still struggle with 3D reasoning tasks like arranging objects in space according to open-ended language instructions, particularly in dense and physically constrained environments. We introduce \method, a framework and scene layout representation that exploits the semantic knowledge of Vision-Language Models (VLMs) and supports differentiable optimization to ensure physical plausibility. \method employs VLMs to generate two mutually reinforcing representations from visually marked images, and a self-consistent decoding process to improve VLMs spatial planning. Our experiments show that \method addresses the limitations of existing LLM and constraint-based approaches, producing physically plausible 3D layouts better aligned with the semantic intent of input language instructions. We also demonstrate that fine-tuning VLMs with the proposed scene layout representation extracted from existing scene datasets can improve their reasoning performance.\looseness=-1

%Additionally, we introduce a self-consistency decoding procedure to 
%for  from visually marked images of the existing scene, 
%\method uses to maintain consistency across multiple abstraction levels.
% To evaluate our framework, we introduce the \textit{Detailed Scene Layout Benchmark}, designed for testing complex indoor scenes with high object density and rich semantic content. 

\end{abstract}    
%\vspace{-0.3cm}
\section{Introduction}
\label{sec:intro}

\begin{table*}[t]
    \centering
    %\resizebox{\textwidth}{!}{%
    \begin{tabular}{lll}
    \toprule
         Type & Notation & Explanation\\
         \midrule
         Positional & $\mathcal{L}_{\text{distance}}(p_i, p_j, d_{\text{min}}, d_{\text{max}})$ & The distance between the two assets should fall within the range $\left[d_{\text{min}}, d_{\text{max}}\right]$ . \\
         Positional & $\mathcal{L}_{\text{on\_top\_of}}(p_i, p_j, b_i, b_j)$ & Position one asset on top of another. \\
         Rotational & $\mathcal{L}_{\text{align\_with}}(p_i, p_j, \phi$) & Align two assets at a specified angle $\phi$. \\
         Rotational & $\mathcal{L}_{\text{point\_towards}}(p_i, p_j, \phi$) & Orient one asset to face another with a offset angle $\phi$. \\
         Mix & $\mathcal{L}_{\text{against\_wall}}(p_i, w_j, b_i$) & Place an asset again wall $w_j$.\\
         \bottomrule
    \end{tabular}
    \caption{\textbf{Spatial Relation Definition}. We define five spatial relations that determine how objects are placed relative to each other and within the room. Each constraint is associated with a differentiable cost function based on the object poses, which can be used for optimization. Our spatial relations related to rotation do not require a fixed reference frame and instead are based on relative poses between objects. %\nh{Not sure what work the variables column is doing. Maybe give formulas?}
    }
    \vspace{-0.2cm}
    \label{tab:grammar_short}
    %}
\end{table*}
%%%%%%%%%%%%%%%%%%%%%%%%%%%%%%%%%%%% version 2
Spatial reasoning and planning involve understanding, arranging, and manipulating objects in 3D space within the constraints of the physical world. These skills are essential for autonomous agents to navigate, plan tasks, and physically interact with objects in complex environments. Automatically generating diverse, realistic scenes in simulation has become crucial for scaling up data to train autonomous agents with enhanced spatial reasoning abilities. In this paper, we advance this goal by addressing open-universe layout generation, which involves creating diverse layouts based on unlabeled 3D assets and free-form language instructions.

Traditional scene synthesis and layout generation methods are often constrained by predefined object categories and patterns of object placements learned from synthetic scene datasets~\cite{wei2023lego, nauata2021house, paschalidou2021atiss}, preventing them from achieving the diversity of scene layouts seen in the real world. Recent methods leverage Large Language Models (LLMs) for open-universe layout generation by utilizing the spatial commonsense embedded in language and program code. However, a key challenge is to achieve both physical plausibility and semantic coherence, as illustrated in Fig.~\ref{fig:pull}. Methods that predict numerical object poses (e.g., LayoutGPT~\cite{feng2024layoutgpt}) often produce layouts with object collisions or out-of-bound placements. Other methods, such as Holodeck~\cite{yang2024holodeck}, attempt to improve physical plausibility by predicting spatial relations between assets and solving a constraint optimization problem. However, these approaches either struggle to find feasible solutions for scenes with large numbers of objects or output layouts that lack the semantic nuances specified in the language instructions.

%In this paper, we introduce \method, an open-universe layout generation method that effectively achieves both physical plausibility and semantic alignment. Our approach is based on the insight that numerical object poses and spatial relations are complementary representations, and their combination through differentiable optimization enables robust layout generation.
% A set of differentiable objectives is also introduced alongside the set of spatial relations, with one-to-one correspondence.
% The set of spatial relations we introduce  differentiable objectives. 
%Thus, we introduce spatial relations and their corresponding differentiable objectives.
%More specifically, our method predicts both representations, with the numerical value prediction as initialization to the optimization process. The set of spatial relations, introduced alongside their corresponding differentiable objectives, are then used as objective functions to maintain semantic coherence while the layout is differentiably optimized to ensure physical plausibility. These objectives can be combined with other physics-related objectives to refine object placements while preserving the estimated layout. Our scene representation also leverages the VLM’s spatial planning capabilities through visually marked images, allowing accurate placement of grouped objects within the scene. The two different abstractions of numerical values and spatial relations further improve spatial reasoning, regulated through self-consistent decoding.

In this paper, we introduce \method, an open-universe layout generation method that effectively achieves both physical plausibility and semantic alignment. Our approach leverages the complementary nature of numerical object poses and spatial relations, combining them within a differentiable optimization framework to enable robust layout generation. More specifically, \method first predicts numerical object poses as initialization for the optimization process. Then, \method jointly optimizes for physics-based objectives and spatial relations, each with their corresponding differentiable objectives. The physics-based objectives ensure physical plausibility, while the spatial relations preserve the overall layout semantics during the optimization process. To improve VLM capabilities for spatial grounding, \method uses visually marked images, allowing accurate estimation of object placements within the scene, especially when existing objects exist. We also introduce a self-consistency decoding process that allows \method to focus on the semantically meaningful spatial relations during optimization. The synergy between numerical values and spatial relations, regulated through self-consistent decoding, ensures accurate and coherent scene layouts.

Our contributions are as follows: first, we introduce a novel scene layout representation that can be combined with differentiable optimization to generate diverse layouts. The scene representation builds on two complementary representations—numerical pose estimates and spatial relations with matching differentiable objectives. Second, we show that we can use VLMs and a self-consistency decoding process to generate our scene layout representation using visually marked scene and asset renderings. Third, through systematic evaluation across 11 room types, we achieved significant improvements when compared to the current best-performing method. Fourth, we show that fine-tuning open-source models on our scene representation with synthetic data yields substantial performance improvements, even for models that struggle with 3D layout generation.

\section{Related Work}
\label{sec:formatting}

% Refer to the following survey:
%https://docs.google.com/spreadsheets/d/1DpdvVVkicc9vAHkiLZDM9QXOq23Z_1_S/edit?usp=sharing&ouid=115876959712993195873&rtpof=true&sd=true 

\subsection{Layout Generation \& Indoor Scene Synthesis}
%Refer to the open-vocabulary layout generation section in our Literature review
%Recently, advances in indoor scene synthesis 
%Layout generation mainly falls into two lines of research: 1) generating 3D scenes directly from textual descriptions ~\citep{schult2024controlroom3d, zhou2024gala3d, po2024compositional, epstein2024disentangled, fang2023ctrl} and 2) generating textual or graph representations that can be converted into 3D scenes with object retrieval\citep{rahamim2024lay, feng2024layoutgpt, fu2025anyhome, yang2024holodeck, lin2024instructscene, ocal2024sceneteller, yang2024physcene, cheng2024legent, chang1703sceneseer, wang2024chat2layout, tam2024scenemotifcoder, yang2024llplace, aguina2024open, ccelen2024design, rahamim2024lay}. In the first line of research, they typically generate 3D representations of the scene by Neural Radiance Field or 3D Gaussian Splatting given arbitrary text descriptions. However, the outputs of these methods are often point clouds, unstructured meshes, or isosurfaces extracted from density fields, which cannot be readily used for robotics applications. %They also lack the flexibility to accomodate to new objects or room types. 

Recent advances in indoor scene synthesis have explored two main directions. One line of work leverages the strong generative priors of image generation models, often using Neural Radiance Fields (NeRFs) or Gaussian splats as the output representation~\citep{schult2024controlroom3d, zhou2024gala3d, po2024compositional, epstein2024disentangled}. However, these generated scenes lack separable, manipulable objects and surfaces, rendering them unsuitable for robotics applications where precise object interactions are required.
%However, the generated outputs are often unstructured: point clouds, implicit surfaces, or isosurfaces extracted from density fields. This lack of separable, manipulable objects makes these representations unsuitable for robotics applications, where precise object interactions are required.
Another line of research focuses on generating scenes using intermediate representations, such as scene graphs or scene layouts, combined with an asset repository~\citep{rahamim2024lay, feng2024layoutgpt, fu2025anyhome, yang2024holodeck, lin2024instructscene}. %Many works focus on learning the distribution of scenes from a dataset~\cite{paschalidou2021atiss,wei2023lego}, but 
The advent of Large Multimodal Models (LMMs) has enabled open-vocabulary 3D scene synthesis, supporting the flexible generation of scenes without dependence on predefined labels or categories~\cite{ccelen2024design,aguina2024open}. For example, LayoutGPT~\cite{feng2024layoutgpt} prompt language models to directly generate 3D Layouts for indoor scenes. Holodeck~\citep{yang2024holodeck} uses LLMs to generate spatial scene graphs and then uses the specified scene graph to optimize object placements. \method also falls into this line of work, but instead of using LLMs, we generate scene layout representations from image and text inputs using VLMs. Additionally, we introduce a differential optimization process as opposed to solving a constraint satisfaction problem with search~\cite{yang2024holodeck}.

%The Recent advances in have enabled  Early approaches in scene synthesis utilized LLMs to encode textual input into vector representations that guided object placement within scenes [30, 31, 48, 53]. ~\citet{feng2024layoutgpt} extended this work with  This approach functions similarly to a retrieval system, positioning objects based on absolute coordinates. Building on this, ~\citet{fu2025anyhome} introduced a dataset-free, open-vocabulary framework for 3D home layouts, pushing towards adaptable, user-driven configurations. , optimizing object placements to produce more coherent and diverse scenes for embodied AI applications. It aims to mitigate some of the challenges seen in earlier 3D scene synthesis approaches, specifically by using spatial relational constraints rather than direct coordinate placements, which helps reduce issues like object overlap and out-of-bounds placements. However, these methods often face challenges due to the geometric limitations of GPT models, leading to issues such as object overlap or out-of-bound placements.

\begin{figure}[t]
    \centering
    \includegraphics[width=\linewidth]{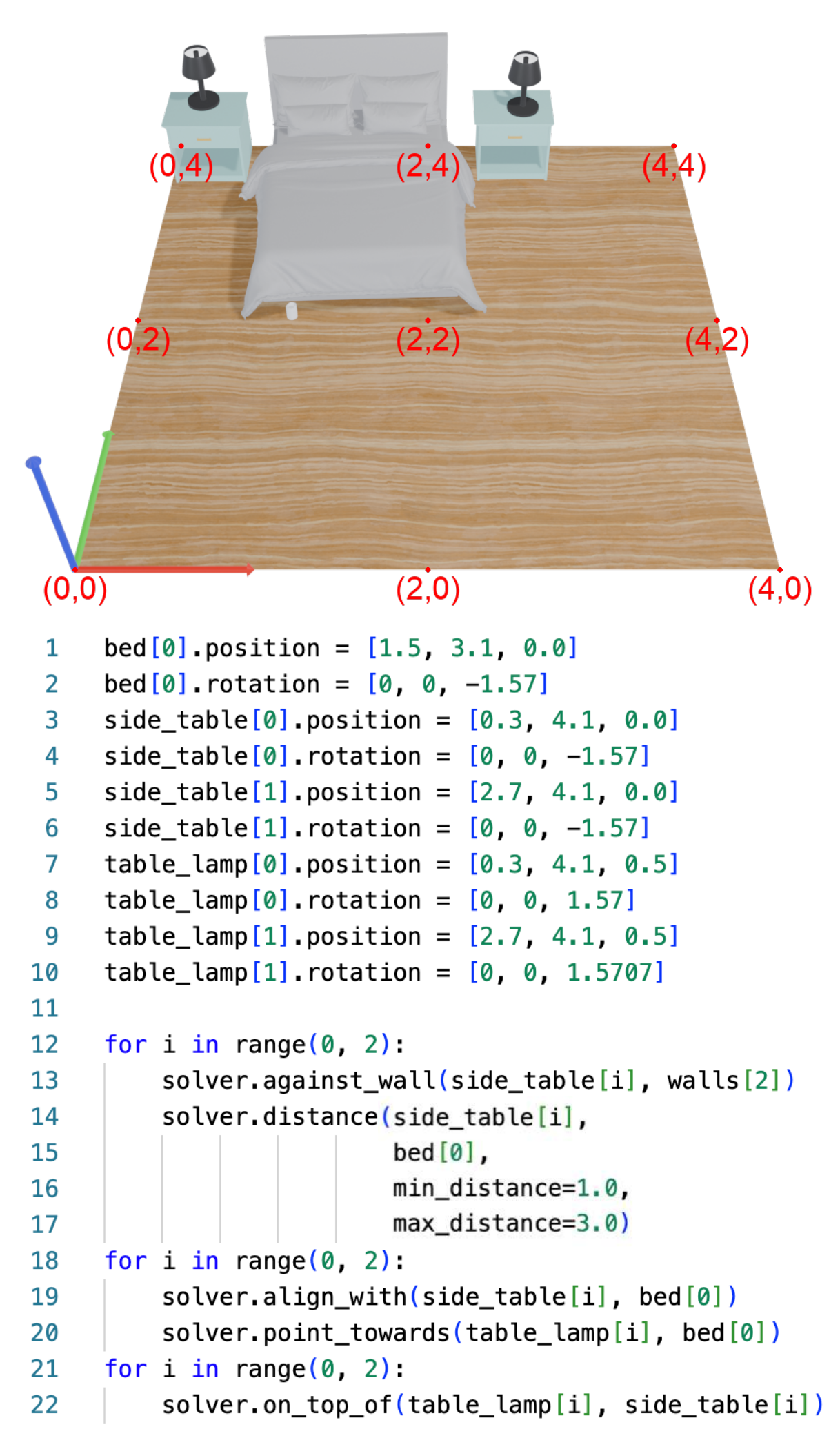}
    \vspace{-0.5cm}
    \caption{\textbf{Example Scene Representation}. Example of our scene representation for a bedroom. Our scene representation consists of numerical estimates of object poses and spatial relations corresponding to objective functions on these poses. Having the VLMs generate the initial estimates allows us to exploit the semantic knowledge in the large models, and having spatial relations amenable to optimization allows us to generate physically precise placements.}%  This scene representation is \textit{semantically expressive}—capable of modeling the diverse relations between different furniture pieces in this bedroom —and \textit{geometrically precise}, ensuring physically plausible object placement.}
    % \nh{If you need to save space, image has a lot of bare room space}}
    \label{fig:scene_representation_example}
    %\vspace{-0.7cm}
\end{figure}

\subsection{Vision-Language Models for 3D Reasoning}

Recent works have explored the spatial reasoning capabilities of vision-language models (VLMs). Some studies have trained 3D visual encoders on representations like point clouds and meshes to improve tasks such as 3D scene understanding, question answering, navigation, and planning~\cite{hong20233d,3dgrand,scenellm,ll3da,gpt4point,pointllm,chatscene}. Other research has adapted 2D VLMs for spatial reasoning by fine-tuning them on visual question-answering datasets involving metric and qualitative spatial relations grounded in 3D environments~\cite{chen2024spatialvlm,cheng2024spatialrgpt}. A related direction is to reconstruct the 3D scene based on 2D images by training on large synthetic data~\cite{scenescript}. However, these approaches primarily focus on perception tasks and do not extend to generating 3D structures. In contrast, our work uses 2D VLMs for the task of 3D layout generation, utilizing techniques originally developed for visual reasoning—such as leveraging images of a scene from different viewing angles~\cite{agent3d} and visual markers~\cite{yang2023set,vip,coordinate}—repurposed here for spatial planning. We also investigate fine-tuning VLMs for 3D layout generation tasks and observe significant improvements in the performance of open-source models.

\section{Problem Formulation}
%\vspace{-0.2cm}
\label{sec:formulation}

The problem of open-universe 3D layout generation is to arrange any assets within a 3D environment based on natural language instructions. Formally, given a layout criterion \( \ell_{\text{layout}} \) in natural language, a space defined by four walls oriented along the cardinal directions $\{w_1, \dots, w_4\}$, and a set of \( N \) 3D meshes $\{m_1, \dots, m_N\}$, the goal is to create a 3D scene that faithfully represents the provided textual description. Following prior work~\cite{yang2024holodeck,aguina2024open}, we assume that the input 3D objects are upright and an off-the-shelf vision-language model (VLM) (i.e., GPT-4o~\cite{achiam2023gpt}) is employed to determine their front-facing orientations. The VLM also annotates each object with a short textual description \( s_i \), and the dimensions of its axis-aligned bounding box after rotating to face the \( +x \) axis will be represented as \( b_i \in \mathbb{R}^3 \). The target output of layout generation is each object's pose \(p_i = (x_i, y_i, z_i, \theta_i)\), including the object's 3D position and rotation about the \( z \)-axis.
\section{\method} \begin{figure*}[!ht]
    \centering
    \includegraphics[width=\textwidth]{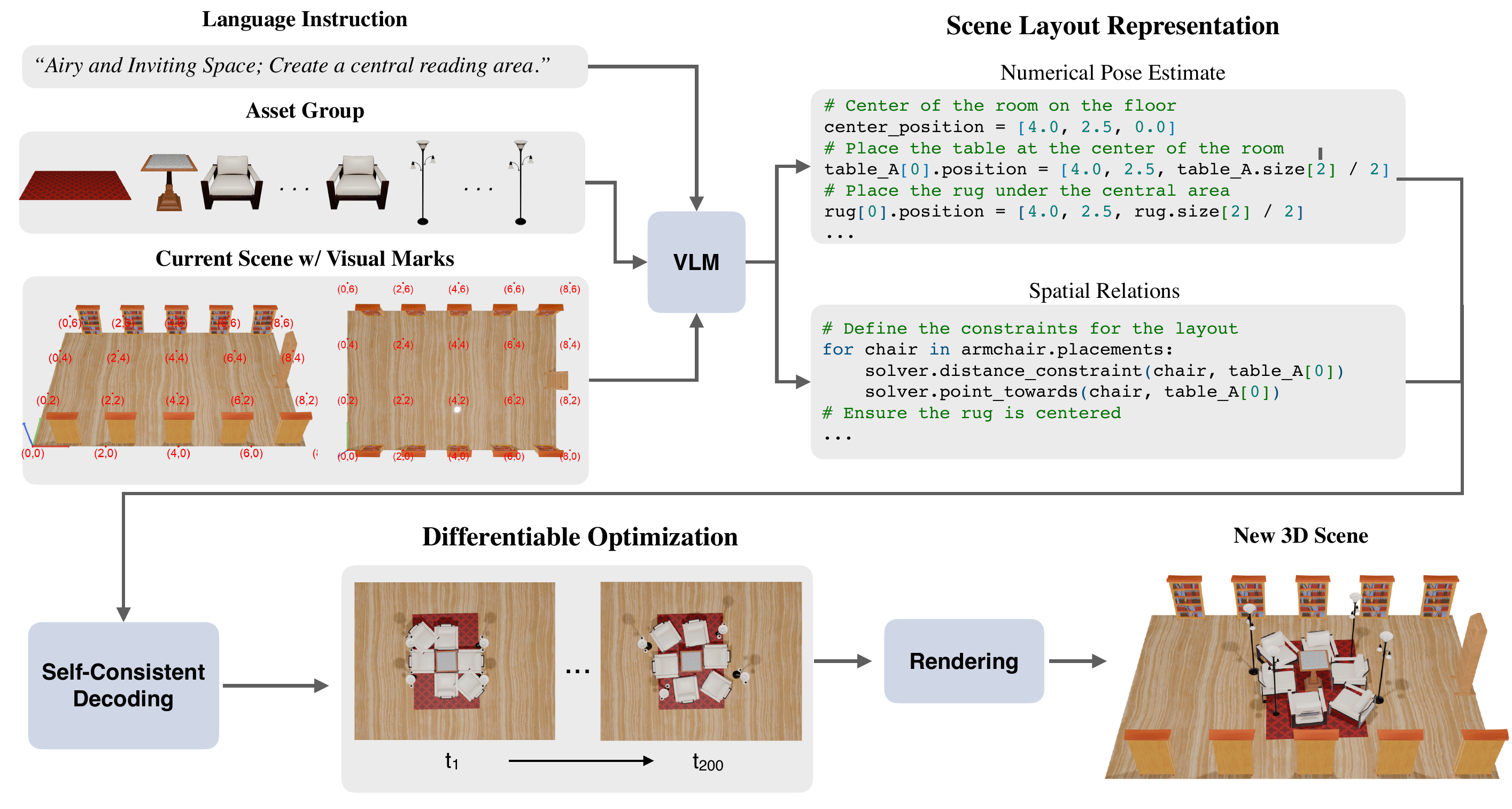}
    \vspace{-0.5cm}
    \caption{\textbf{\method}. We illustrate the proposed process of generating 3D scene layout with Vision-Language Models.}
    \label{fig:main}
   % \vspace{-0.5cm}
\end{figure*}

%\vspace{-0.2cm}
In this paper, we propose \method, a method to generate physically plausible arrangements of unlabeled 3D assets based on open-vocabulary natural language descriptions of the desired layouts outlined in Figure~\ref{fig:main}. Our approach employs vision-language models (VLMs) to generate code for our proposed scene layout representation that specifies both an initial layout $\{\hat{p}_i\}_{i=1}^{N}$, as well as a set of spatial relations between assets (and walls). This representation is then used to produce the final object placements through optimization:
\begin{align*}
\argmin_{\{p_i\}_{i=1}^{N}} \left(\mathcal{L}_{\text{semantic}} + \mathcal{L}_{\text{physics}}\right), \\
\; \text{initial solution } \{\hat{p}_i\}_{i=1}^{N}.
%& \text{with initial solution}\, \{\hat{p}_i\}_{i=1}^{N},
\end{align*}
where $\mathcal{L}_{\text{semantic}}$ is the objective function decoded from the scene layout representation and $\mathcal{L}_{\text{physics}}$ is the objective function we employ to ensure physical plausibility.

In the following sections, we first introduce our scene layout representation, explain how VLMs can reliably generate this representation using visual marks, elaborate on our proposed \textit{self-consistent decoding} process alongside the differentiable optimization process, and discuss finetuning VLMs to improve their understanding of our representation for improved performance.

%This scene representation includes numerical estimates of object poses and spatial relations that serve as constraints. The numerical poses serve as the initialization, and the constraints allow a differentiable optimization solver to refine the layout, ensuring it is both physically valid and semantically aligned with the language instructions. 

\subsection{Scene Layout Representation}
To generate layouts from open-ended language instructions and 3D assets, a desirable representation must be \textit{semantically expressive} for diverse language specifications and \textit{physically precise} to ensure plausible 3D layouts. Our representation includes (a) numerical estimates of object poses $\{\hat{p}_i\}_{i=1}^{N}$ and (b) spatial relations with differentiable objectives. Initial estimates provide a starting point for optimization, while differentiable objectives guide the process, addressing the challenge of directly predicting physically plausible layouts.  
% This is crucial because predicting final object poses that are both physically and geometrically plausible is inherently challenging, especially in complex scenes. 
The initial layout is key in the optimization process, as poor initialization can lead to a suboptimal layout (e.g., a table separating the room into two halves where each chair is placed on opposite sides, when the language instruction instructs them to be placed on the same side). The spatial relations are crucial to ensure that the layout semantics are maintained while the layout is adjusted for physical plausibility.
% At the same time, differentiable objectives are also important, as they introduce constraints that guide the optimization process, addressing the inherent challenge of predicting geometrically plausible layouts directly. 
% These objectives being continuously differentiable is particularly valuable in complex scenarios, where the intricacies of spatial relationships make it difficult to rely solely on hard constraints, as they may overly restrict the solution space and hinder the discovery of feasible configurations. 
In Figure~\ref{fig:scene_representation_example}, we show an example of our scene representation. 

\myparagraph{Differentiable Spatial Relations} 
The goal of these spatial relations is twofold: (a) to capture the semantics of the input language instruction and (b) to preserve these semantics during optimization for physical plausibility. For example, consider the instruction, ``set up a dining table." A vision-language model might initially generate poses where the chairs overlap with the table. Our differentiable spatial relations are designed to adjust these poses during optimization—preventing overlaps—while maintaining the essential semantics, such as ``chairs should be positioned near the dining table in a dining setup." To design a set of spatial relations that can capture a wide range of semantics for indoor scenes, we devise five expressive spatial relations: two positional objectives (i.e., \textit{distance}, \textit{on\_top\_of}), two orientational objectives (i.e., \textit{align\_with}, \textit{point\_towards}), and one wall-related objective (i.e., \textit{against\_wall}) that pertains to both the position and orientation of an asset. Note that our spatial relations do not rely on a fixed reference frame, unlike classic spatial relations like \textit{in\_front\_of} and \textit{left\_of}. Each spatial relation corresponds to a differentiable objective function with optional parameters that VLMs can choose to impose on object poses; for example, a \textit{distance} imposes a higher loss if objects are outside a specified distance range, and the VLMs can decide the lower and upper bound for this function.
%The goal of the spatial relations is to (a) capture the semantics of a layout and the input language instruction and to ensure that Our representation builds on a set of spatial relations that govern how objects are positioned and oriented relative to each other. We devise five spatial relations: two positional objectives(i.e., \textit{distance}, \textit{on\_top\_of}), two orientational objectives (i.e., \textit{align\_with}, \textit{point\_towards}), and one wall-related objective (i.e., \textit{against\_wall}) that pertains to both the position and orientation of an asset. %a combination of both to constrain the placement of objects relative to each other and the room. 
Table~\ref{tab:grammar_short} presents the notations and meanings of these spatial relations. Formally, we denote a set of spatial relations derived from a scene layout representation as $\mathcal{R}$.
%\[
% = \left\{ \mathcal{L}_{k} \mid \mathcal{L}_{k} \in \mathcal{T}, \; k = 1, 2, \dots, M \right\}.
%\]
%\[
%\mathcal{T} = \{\mathcal{L}_{\text{distance}}, \mathcal{L}_{\text{on\_top\_of}}, %\mathcal{L}_{\text{align\_with}}, \mathcal{L}_{\text{point\_towards}}, \mathcal{L}_{\text{against\_wall}}\},
%\]
% \[
% \mathcal{R} = \left\{ \mathcal{L}_{k} \mid \mathcal{L}_{k} \in \mathcal{T}, \; k = 1, 2, \dots, M \right\}.
% \]
%\(\mathcal{T}\) is the set of all possible types of spatial relations.
%$\mathcal{R} = \{ \mathcal{L}_{\text{distance}}, \mathcal{L}_{\text{on\_top\_of}}, \mathcal{L}_{\text{align\_with}}, \mathcal{L}_{\text{point\_towards}}, \mathcal{L}_{\text{against\_wall}} \}$. % Empirically, we find that these differentiable objective functions are effective when optimized with gradient descent.
%. The choice of these five specific relations allows us to define a set of expressive constraints that can capture a wide range of semantics for indoor scenes.
%Our orientational relations do not rely on a fixed reference frame, unlike classic spatial relations like \textit{in\_front\_of} and \textit{left\_of}. Each spatial relation corresponds to a differentiable cost function to impose on object poses; for example, a \textit{distance\_constraint} imposes a high cost if objects are outside a specified distance range. % Preliminary experiments indicate that gradient descent is effective for optimization. 
% Refer to Appendix~\label{supp:loss} for the mathematical formulation of the 

\subsection{Generating Scene Layout Representation with Vision-Language Models}
Our approach utilizes the generalization and commonsense reasoning abilities of Vision-Language Models (VLMs) to generate the scene representation outlined above based on the objects, 3D scene, and language instructions. %Previous work has demonstrated that pre-trained language models can both generate numerical values for object poses~\cite{feng2024layoutgpt} and apply common spatial relations from language to constrain object placement~\cite{yang2024holodeck}. 
%Our scene representation leverages these capabilities, requiring the VLM to create a representation with both numerical and relational components based on the provided language and visual inputs. 
To improve the accuracy of the generated scene representation, we employ two techniques: visual prompting with coordinates and self-consistent decoding.

\myparagraph{Visual Prompting} Figure~\ref{fig:main} illustrates our VLM-based scene representation generation process. The VLM’s input includes rendered images of the 3D scene and individual asset views. Prior research has shown that visual cues can improve VLMs' object recognition and spatial reasoning~\cite{yang2023set}. We provide two types of visual annotations for layout generation: coordinate points in the 3D space spaced 2 meters apart to help the VLM gauge dimensions and scale and visualizations of coordinate frames to maintain consistent spatial references. We also annotate the front-facing orientations of each object with directional arrows, which is essential for generating rotational constraints like \textit{aligned\_with} or \textit{point\_towards}. In practice, we first use an LLM to group the assets given the textual descriptions $s_i$ of all the input assets to address context length limitations when handling many 3D assets. Then, we place assets in groups, one group at a time. Before generating each group, we re-render the 3D scene to help the VLM identify unoccupied areas, thus facilitating the physically valid placement of remaining assets.
\begin{figure*}
    \centering
    \includegraphics[width=\textwidth]{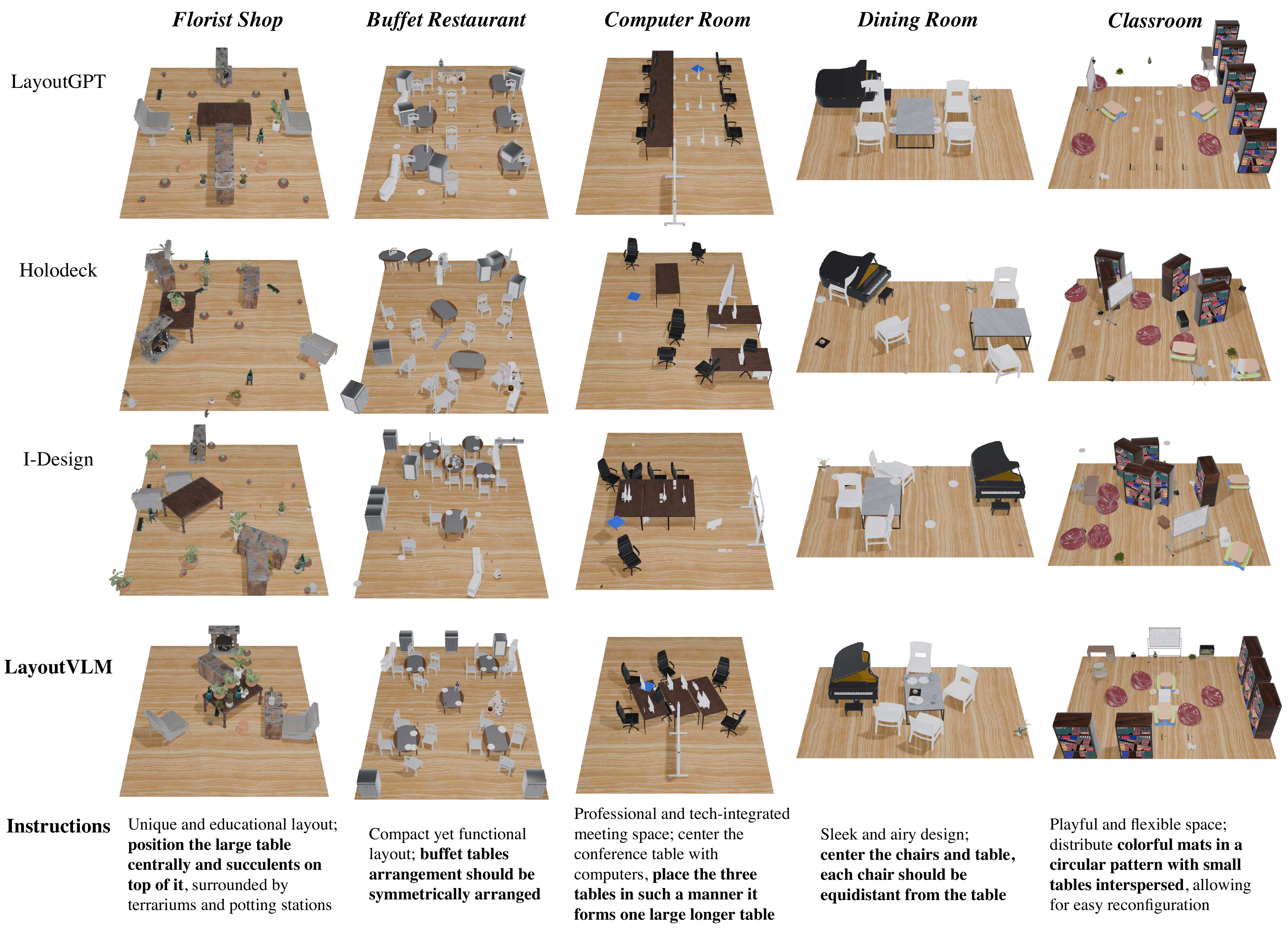}
    %\vspace{-0.5cm}
    \caption{\textbf{Qualitative Comparison.} We compare with baseline methods in generating layouts based on detailed language instructions. Our method is able to generate layouts that closely follow the instructions and adhere to physical constraints.}
    \label{fig:qualitative}
    %\vspace{-0.2cm}
\end{figure*}

\begin{table*}[t]
    \centering\small
    %\vspace{-0.4cm}
    %\scriptsize
    %\setlength{\tabcolsep}{.9pt}
    \resizebox{\linewidth}{!}{%
    \begin{tabular}{l cccc>{\columncolor{lightergray}}c cccc>{\columncolor{lightergray}}c cccc>{\columncolor{lightergray}}c cccc>{\columncolor{lightergray}}c cccc>{\columncolor{lightergray}}c cccc>{\columncolor{lightergray}}c}
    \toprule
    & \multicolumn{5}{c}{\textbf{Bedroom}} & \multicolumn{5}{c}{\textbf{Living Room}} & \multicolumn{5}{c}{\textbf{Dining Room}} & \multicolumn{5}{c}{\textbf{Bookstore}} & \multicolumn{5}{c}{\textbf{Buffet Restaurant}} & \multicolumn{5}{c}{\textbf{Children Room}} \\
    \cmidrule(lr){2-6} \cmidrule(lr){7-11} \cmidrule(lr){12-16} \cmidrule(lr){17-21} \cmidrule(lr){22-26} \cmidrule(lr){27-31}
    \textbf{Methods} & CF & IB & Pos. & Rot. & PSA & CF & IB & Pos. & Rot. & PSA & CF & IB & Pos. & Rot. & PSA & CF & IB & Pos. & Rot. & PSA & CF & IB & Pos. & Rot. & PSA & CF & IB & Pos. & Rot. & PSA \\
    \midrule
    LayoutGPT & 100.0 &  66.7 &  85.7 &  85.9 &  52.2 &  44.4 &  11.1 &  74.7 &  64.4 &   \textbf{9.6} &  88.9 &  22.2 &  76.0 &  68.9 &  14.8 &  88.9 &  55.6 &  80.9 &  79.4 &  35.9 & 100.0 &  33.3 &  81.2 &  83.3 &  26.9 & 100.0 &   0.0 &  80.9 &  82.6 &   0.0  \\
    Holodeck &  88.9 &  22.2 &  69.3 &  67.9 &  14.1 &  77.8 &   0.0 &  66.3 &  55.6 &   0.0 &  88.9 &   0.0 &  38.0 &  36.6 &   0.0 &  55.6 &   0.0 &  65.7 &  59.0 &   0.0 &  77.8 &  11.1 &  47.7 &  42.4 &   7.4 &  77.8 &  22.2 &  72.7 &  70.0 &  18.7 \\
    I-Design & 100.0 & 77.8 & 72.1 & 65.4 & 51.5 & 33.3 & 11.1 & 62.6 & 46.7 & 0.0 & 88.9 & 66.7 & 76.4 & 66.4 & 34.8 & 66.7 & 11.1 & 68.1 & 69.4 & 5.2 & 100.0 & 55.6 & 63.5 & 57.1 & 35.2 & 77.8 & 55.6 & 78.1 & 75.1 & 34.8 \\
    \textbf{\method} & 88.9 & 100.0 &  82.3 &  74.9 & \textbf{68.8} &  22.2 &  77.8 &  68.6 &  54.4 &  \textbf{9.6} & 88.9 & 100.0 &  63.4 &  56.9 &  \textbf{51.1} & 55.6 & 100.0 &  82.0 &  82.8 &  \textbf{49.8} & 88.9 &  88.9 &  74.3 &  64.8 &  \textbf{51.5} & 100.0 & 100.0 &  81.9 &  88.2 &  \textbf{88.5}\\
    \bottomrule
    \end{tabular}
    }
    \resizebox{\linewidth}{!}{%
    \begin{tabular}{l cccc>{\columncolor{lightergray}}c cccc>{\columncolor{lightergray}}c cccc>{\columncolor{lightergray}}c cccc>{\columncolor{lightergray}}c cccc>{\columncolor{lightergray}}c |cccc>{\columncolor{lightergray}}c}
    \toprule
    & \multicolumn{5}{c}{\textbf{Classroom}} & \multicolumn{5}{c}{\textbf{Computer Room}} & \multicolumn{5}{c}{\textbf{Deli}} & \multicolumn{5}{c}{\textbf{Florist Shop}} & \multicolumn{5}{c|}{\textbf{Game Room}} & \multicolumn{5}{c}{\textbf{Average}} \\
    \cmidrule(lr){2-6} \cmidrule(lr){7-11} \cmidrule(lr){12-16} \cmidrule(lr){17-21} \cmidrule(lr){22-26} \cmidrule(lr){27-31}
    \textbf{Methods} & CF & IB & Pos. & Rot. & PSA & CF & IB & Pos. & Rot. & PSA & CF & IB & Pos. & Rot. & PSA & CF & IB & Pos. & Rot. & PSA & CF & IB & Pos. & Rot. & PSA & CF & IB & Pos. & Rot. & PSA \\
    \midrule
    LayoutGPT & 88.9 &   0.0 &  76.3 &  66.7 &   0.0 & 100.0 &  22.2 &  87.8 &  85.2 &  17.8 &  88.9 &   0.0 &  77.2 &  77.9 &   0.0 &  66.7 &  33.3 &  81.6 &  80.2 &  18.3 &  55.6 &  22.2 &  87.0 &  82.9 &   6.7 & \textbf{83.8} &  24.2 &  \textbf{80.8} &  \textbf{78.0} &  16.6\\
    Holodeck & 33.3 &   0.0 &  45.2 &  38.6 &   0.0 & 100.0 &   0.0 &  66.1 &  59.7 &   0.0 &  88.9 &  33.3 &  73.9 &  63.7 &  24.4 &  63.9 &   0.0 &  73.2 &  64.7 &   0.0 &  55.6 &  22.2 &  60.7 &  58.0 &   0.0 & 77.8 & 8.1 & 62.8 & 55.6 & 5.6 \\
    I-Design & 55.6 & 11.1 & 50.7 & 47.0 & 0.0 & 88.9 & 22.2 & 74.0 & 70.7 & 8.9 & 88.9 & 22.2 & 67.8 & 65.9 & 10.4 & 77.8 & 0.0 & 75.5 & 68.3 & 0.0 & 66.7 & 44.4 & 62.8 & 58.9 & 17.0 & 76.8 & 34.3 & 68.3 & 62.8 & 18.0 \\
    \textbf{\method} & 77.8 & 100.0 &  74.6 &  68.6 &  \textbf{48.3} & 100.0 &  88.9 &  85.4 &  84.5 &  \textbf{77.0} & 100.0 &  88.9 &  83.4 &  83.4 &  \textbf{74.6} & 88.9 & 100.0 &  83.4 &  76.4 &  \textbf{68.3} & 88.9 & 100.0 &  73.1 &  70.0 &  \textbf{59.5} & \textbf{81.8} & \textbf{94.9} & \textbf{77.5} & 73.2 & \textbf{58.8}\\
    \bottomrule
    \end{tabular}
    }
    \caption{\textbf{Benchmark Performance.} \method outperformed existing open-universe layout generation methods across 11 room types.}
    \label{tab:benchmark_v3}
    \vspace{-0.2cm}
\end{table*}

\myparagraph{Self-Consistent Decoding} One key challenge is that VLMs struggle with spatial planning; while they may successfully generate spatial relations for pairs of objects, they tend to fail at accounting for overall layout coherence. We hypothesize that self-consistent spatial relations (i.e., "the spatial relations that are also satisfied in the estimated numerical poses of the objects") represent the most critical semantics to preserve during optimization when the object poses are adjusted for better physical plausibility. Thus, we introduce self-consistent decoding for our scene layout representation. Different from standard self-consistency~\cite{wang2022self}, which selects the most consistent answer from multiple reasoning paths following the same format, we require the two distinct but mutually reinforcing representations predicted by the VLM to \textit{self-consistent}. That is, we only retain the spatial relations satisfied with the predicted poses. After self-consistent decoding, we can formally describe the semantic part of the optimization loss as:
\begin{equation}
\label{eq:semantic_loss}
\mathcal{L}_{\text{semantic}} = \sum_{\mathcal{L} \in \mathcal{R}} \mathbbm{1}\left[\mathcal{L}_{\text{i}}(\hat{p}_i, \hat{p}_j, \mathbf{\lambda}) \leq \epsilon\right] \cdot \mathcal{L}_{\text{i}}(p_i, p_j, \mathbf{\lambda}),
\end{equation}
where \(\hat{p}_i\) and \(\hat{p}_j\) are the initial estimated poses, $\mathbf{\lambda}$ represents the additional parameters in the functions (refer to Table~\ref{tab:grammar_short}), and \(\epsilon\) is a threshold value for determining whether the differentiable spatial relation $\mathcal{L}$ is satisfied in the initial estimates.
%To ensure alignment between these two types of output, we assess whether the generated objective functions correspond to the predicted poses. This involves only retaining , improving reliability and accuracy in layout generation. More specifically,
%\[
%\mathcal{L}_{\text{semantic}} = \sum_{(i, j)} \mathbbm{1}\left[\mathcal{L}_{\text{i}}(\hat{p}_i, \hat{p}_j) \leq \epsilon \right] \cdot \mathcal{L}_{\text{i}}(p_i, p_j),
%\]
%where \(\epsilon\) is a small threshold for constraint satisfaction.
%\sun{Add a simple formula here}

\subsection{Differentiable Optimization}
%\wl{should this be moved to 4.1? Since this is related to how our scene representation can be used to generate a layout rather than using VLM to generate the semantic constraints?}
To generate a scene from the estimated object poses and differentiable constraints, we jointly optimize all objects according to the specified constraints over a set number of iterations. In addition to the spatial relational functions generated by the VLM, we impose Distance-IoU loss on objects' 3D oriented bounding box~\cite{zheng2020distance,zhou2019iou} for collision avoidance: 
\begin{equation}
\label{eq:physics_loss}
\mathcal{L}_{\text{physics}} = \sum_{i=1}^{N} \sum_{\substack{j=1 \\ j \neq i}}^{N} \mathcal{L}_{\text{DIoU}}(p_i, p_j, b_i, b_j).
\end{equation}
%For collision avoidance, we compute intersections between 3D bounding boxes, assigning high costs to any overlapping configurations. An out-of-bound constraint further restricts objects within the scene boundaries, penalizing any part that extends beyond the limits. 
%This gives us the final loss function 
Eq.~\ref{eq:semantic_loss} and Eq.~\ref{eq:physics_loss} form the final objective function for the optimization problem. We use projected gradient descent (PGD) to optimize for this objective, projecting assets within the physical boundary every fixed number of iterations during the optimization. With the objective function, we can confine objects within the boundary and avoid unwanted intersections, ensuring physically valid layouts.

\subsection{Finetuning VLMs with Scene Datasets}
Our scene representation can model a wide variety of physically valid and semantically meaningful 3D layouts. Additionally, we can fine-tune VLMs to quickly adapt to this representation, enabling the generation of specific types of layouts. This scene representation can be automatically extracted from scene layout datasets without requiring manual annotations. Specifically, given a set of posed objects in a 3D scene, we apply the preprocessing procedure outlined in Section~\ref{sec:formulation} to obtain both textual descriptions and oriented bounding boxes for each object. After canonicalizing the objects, we compute cost values for our defined spatial relations based on the ground-truth positions and orientations of the objects, using heuristic thresholds to determine whether each spatial relation is satisfied. The resulting scene representation includes both raw object poses and the satisfied spatial relations, which we then use to fine-tune VLMs to generate these scene representations from input objects and scene renderings. In our implementation, we use the 3D-Front dataset to extract training data for around 9000 rooms. Our approach is capable of identifying layout patterns in 3D scenes, such as a variable number of chairs around a table, nightstands positioned beside a bed, or an entertainment area comprising a TV, coffee table, and sofa. We investigate fine-tuning two VLMs for the layout generation task: the closed-source \textit{GPT-4o}~\cite{achiam2023gpt} and the open-source \textit{LLaVA-NeXT-Interleave}~\cite{li2024llava}.

\begin{figure*}[t]
    \centering
    \includegraphics[width=\textwidth]{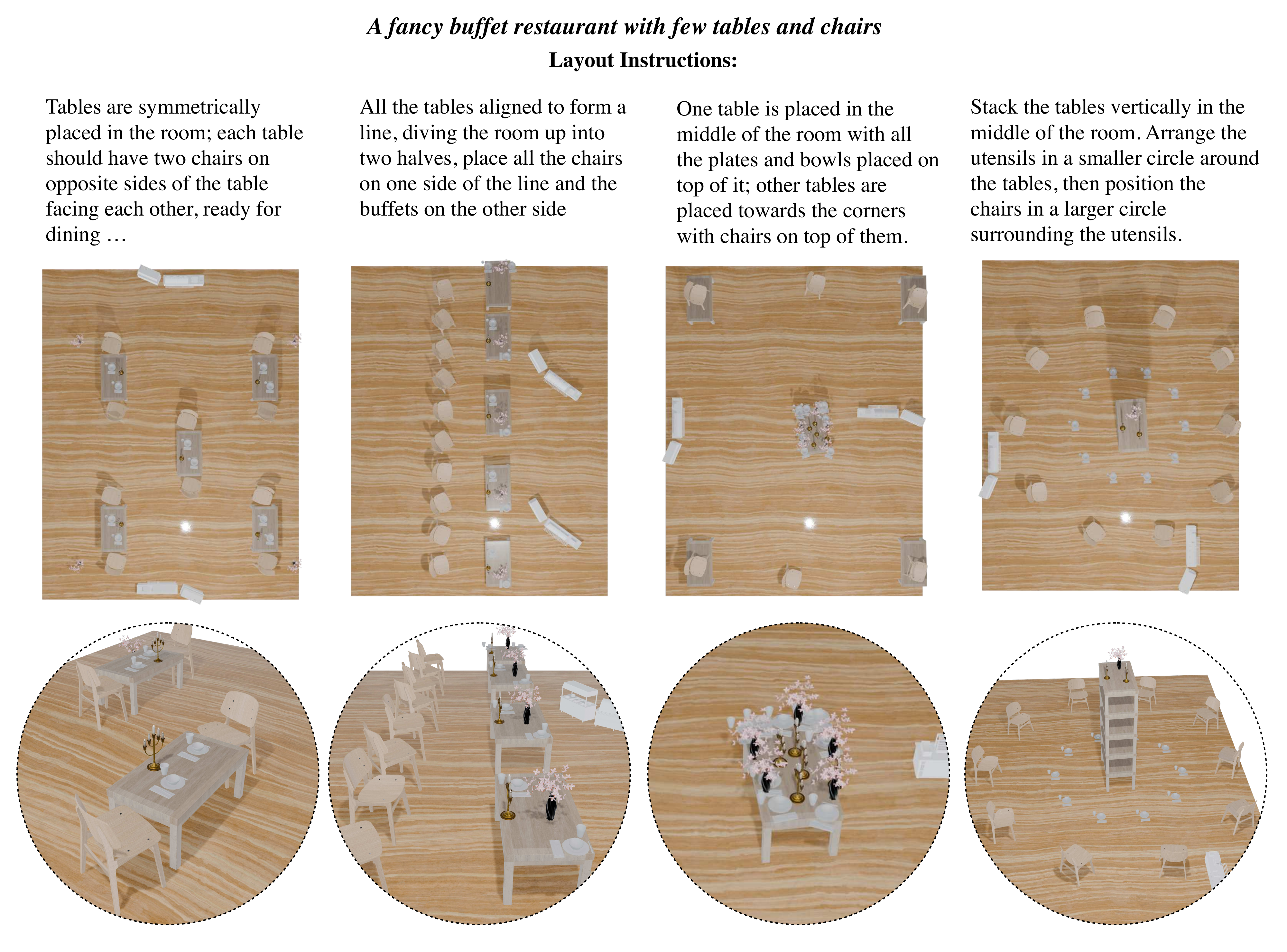}
    \caption{\textbf{Examples of Following Detailed Instructions.} We show the same set of assets arranged with different language instructions. The latter two examples show that \method can closely follow the prompts even when the desired layouts are unconventional.}
    \label{fig:multi_layout}
\end{figure*}
\vspace{3mm}
\section{Experiments}
%\vspace{-1mm}
%1. to evaluate open-vocabulary 3d layout generation, we propose a new benchmark \textbf{Cluttered Scene} ...
%talk about the benchmark \sun{talk about the shortcomings of existing benchmark like 3D-FRONT or smth else}
%\sun{talk about how existing datasets that are commonly used in existing literature are mostly limited to a set of object category and residential room types mostly evaluate scene synthesis }
% In this section, we first explain why existing datasets are not suitable for open-universe layout generation and introduce the new benchmark ... 
% Then we design our experiments to answer three questions:
% \textbf{Q1:} Does \method outperform existing methods on open-universe 3D layout generation?
% \textbf{Q2:} Do the design choices we made about  help with \method's performance 
% \textbf{Q3:} In cases where we have some 3D dataset of ... , Can we finetune ...
In our experiments, we aim to answer the following three questions:
\textbf{Q1:} Does \method outperform existing methods on open-universe 3D layout generation?
\textbf{Q2:} Are the proposed scene layout representation and VLM-based framework essential for generating prompt-aligned 3D layouts that are physically plausible?
\textbf{Q3:} With limited data, can we improve upon pre-trained VLMs' ability to generate the proposed scene layout representation?\looseness=-1
% Is our scene layout representation more effective than %to better generascene layout representation when being fine-tuned on limited data and generalize to new 3D assets? 

%\textbf{Q3:} Can pretrained VLMs easily adapt to our scene layout representation when being fine-tuned on limited data and generalize to new 3D assets? 

\subsection{Experimental Setup}
%\subsection{Experiment Setup}
%\sun{a lot of works assume you can just "reject" assets, }
%\subsection{Benchmarking 3D Layout Generation}
%\sun{mention that we focus on indoor scenes in this paper}
% 3D-FRONT~\cite{fu20213d} is a dataset commonly used to evaluate indoor scene synthesis. However, it lacks the diversity and complexity of realistic and densely populated scenes. More specifically, the assets are limited to a set of fixed categories, and the layouts are mostly limited to ``regular'' arrangements designed by professional interior designers.

\myparagraph{Evaluation} In our evaluation, we assess models’ abilities in 1) open-vocabulary language instructions, 2) predicting layouts for objects beyond predefined categories, and 3) generating accurate object placements within boundaries while avoiding collisions. We create test cases across 11 room types, with three rooms per type and up to 80 assets per room. Each test case includes human-verified, pre-processed 3D assets sourced from Objaverse ~\cite{deitke2023objaverse}, language instructions generated by GPT-4, and a floor plan defining space dimensions. All methods use the same pre-processed assets. Details on test case generation are in the appendix.

\myparagraph{Evaluation Metrics} 
We evaluate generated 3D layouts on physical plausibility and semantic coherence with respect to the provided language instructions. Physical plausibility is measured using the \textit{Collision-Free Score (CF)} and \textit{In-Boundary Score (IB)}. All assets are enforced to be placed, with remaining assets randomly placed if a method fails. Semantic coherence is assessed using \textit{Positional Coherency (Pos.)} and \textit{Rotational Coherency (Rot.)}, measuring alignment with the input prompt. To evaluate semantic coherence across layouts without groundtruth, we use \textit{GPT-4o} to score layouts based on top-down and side-view renderings and the language instructions, leveraging its effectiveness as a human-aligned evaluator in text-to-3D generation~\cite{wu2024gpt}. We also introduce the \textit{Physically-Grounded Semantic Alignment Score (PSA)} to assess physical plausibility and semantic alignment, assigning 0 if assets cannot be feasibly placed. PSA is calculated simply the GPT-4o rating weighted by physical plausibility. Scores range from 0 to 100, with higher scores indicating better performance.

\myparagraph{Baselines} We evaluate our method against the following baselines: LayoutGPT~\citep{feng2024layoutgpt}, Holodeck~\citep{yang2024holodeck}, and I-Design~\citep{ccelen2024design}, representing existing methods for open-universe layout generation.

% , taking into consideration of whether the scene is physically plausible, therefore providing an overall assessment of both physical plausibility and semantic alignment. Metrics related to physical plausibility are calculated deterministically, whereas all other metrics are evaluated with GPT-4o, ensuring a consistent and reliable assessment.

\subsection{Benchmark Performance}
Our method achieves significantly improved performance over existing methods, as shown in \Cref{tab:benchmark_v3}
. Averaging over 11 room types, it improves the PSA score by $40.8$ compared to the best-performing baseline, I-Design. LayoutGPT generates layouts with high semantic coherence by having LLMs predict precise object placements but often produces physically infeasible layouts. Holodeck struggles to generate valid and accurate scenes due to its search-based approach, which becomes ineffective with the large number of assets and rigid constraints that fail to accommodate diverse language instructions. Notably, our approach significantly reduces objects placed outside room boundaries while maintaining high positional and rotational coherence.
\begin{table}[t]
    \centering%\small
    %\resizebox{\columnwidth}{!}{%
    % \scriptsize
    % \tiny
    %\vspace{-2mm}
    \setlength\tabcolsep{10pt}
    \setlength\extrarowheight{1pt}
    \begin{adjustbox}{width=\linewidth}
    \begin{tabular}{lccc|ccc}
    \hline
     & \multicolumn{3}{c}{User} & \multicolumn{3}{c}{GPT-4o} \\
     \cline{2-4} \cline{5-7}
    %  & \multicolumn{2}{c}{Semantics} & \multicolumn{1}{c}{Overall Rank} & \multicolumn{2}{c}{Semantics} & \multicolumn{1}{c}{Overall Rank} \\
    % \cmidrule(lr){2-3} \cmidrule(lr){4-4} \cmidrule(lr){5-6} \cmidrule(lr){7-7}
     & Position & Rotation & PSA & Position & Rotation & PSA\\
    \hline
    LayoutGPT & 1.91 & 1.86 & 2.77 & 2.00 & 1.82 & 2.83\\
    Holodeck & 3.44 & 3.50 & 3.10 & 3.20 & 3.43 & 3.10 \\
    I-Design & 2.86 & 2.91 & 2.64 & 2.89 & 2.86 & 2.62 \\
    \method &  1.79 & 1.73 & \textbf{1.50} & 1.91 & 1.89 & \textbf{1.45} \\
    \hline
    \end{tabular}
    %}
    \end{adjustbox}
    %\vspace{-8pt}
    \caption{Average ranks based on user ratings and GPT4-o scores.}
    %\vspace{-8pt}
    \label{tab:user_study}
\end{table}

% \begin{table}[t]
%     \centering%\small
%     %\resizebox{\columnwidth}{!}{%
%     % \scriptsize
%     % \tiny
%     \setlength\tabcolsep{10pt}
%     \setlength\extrarowheight{1pt}
%     \begin{adjustbox}{width=\linewidth}
%     \begin{tabular}{lccc|ccc}
%     \hline
%      & \multicolumn{3}{c}{User} & \multicolumn{3}{c}{GPT-4o} \\
%      \cline{2-4} \cline{5-7}
%     %  & \multicolumn{2}{c}{Semantics} & \multicolumn{1}{c}{Overall Rank} & \multicolumn{2}{c}{Semantics} & \multicolumn{1}{c}{Overall Rank} \\
%     % \cmidrule(lr){2-3} \cmidrule(lr){4-4} \cmidrule(lr){5-6} \cmidrule(lr){7-7}
%      & Position & Rotation & PSA & Position & Rotation & PSA\\
%     \hline
%     LayoutGPT & 1.79 & 1.77 & 2.74 & 1.97 & 1.84 & 2.88\\
%     Holodeck & 3.37 & 3.39 & 3.07 & 3.11 & 3.34 & 2.96 \\
%     I-Design & 2.94 & 3.04 & 2.71 & 2.90 & 2.81 & 2.67 \\
%     \method &  1.90 & 1.80 & \textbf{1.48} & 2.01 & 2.01 & \textbf{1.49} \\
%     \hline
%     \end{tabular}
%     %}
%     \end{adjustbox}
%     \vspace{-8pt}
%     \caption{Average ranks based on user ratings and GPT4-o scores.}
%     \vspace{-3pt}
%     \label{tab:user_study}
% \end{table}
%\begin{wraptable}{r}{0.56\linewidth} % Adjust 'r' for right-aligned, 'l' for left
\begin{table}[t]
    %\vspace{-1mm}
    \centering
    \setlength\tabcolsep{10pt}
    \setlength\extrarowheight{1pt}
    \begin{adjustbox}{width=.8\linewidth}
        \begin{tabular}{lccc}
            \toprule
             & Position & Rotation & PSA \\
            \midrule
            Within Users & 0.51 & 0.57 & 0.50 \\
            Users With GP4-o & 0.49 & 0.61 & 0.46\\
            \bottomrule
        \end{tabular}
    \end{adjustbox}
    \caption{User-User and User-GPT4o Alignment/Agreement}
    %\vspace{-4pt}
    \label{tab:alignment}
\end{table}
%\end{wraptable}

% \begin{wraptable}{r}{0.5\linewidth} % Adjust 'r' for right-aligned, 'l' for left
%     \centering
%     \setlength\tabcolsep{10pt}
%     \setlength\extrarowheight{1pt}
%     \begin{adjustbox}{width=\linewidth}
%         \begin{tabular}{lccc}
%             \toprule
%              & Position & Rotation & PSA \\
%             \midrule
%             Within & 0.54 & 0.63 & 0.52 \\
%             With GP4-o & 0.56 & 0.69 & 0.60\\
%             \bottomrule
%         \end{tabular}
%     \end{adjustbox}
%     \caption{User-User and User-GPT4o Alignment/Agreement}
%     \vspace{-3pt}
%     \label{tab:alignment}
% \end{wraptable}
\begin{figure}[t]
    \centering
    \vspace{-15pt}
    \begin{subfigure}{0.45\linewidth}
        \centering
        \includegraphics[width=\textwidth]{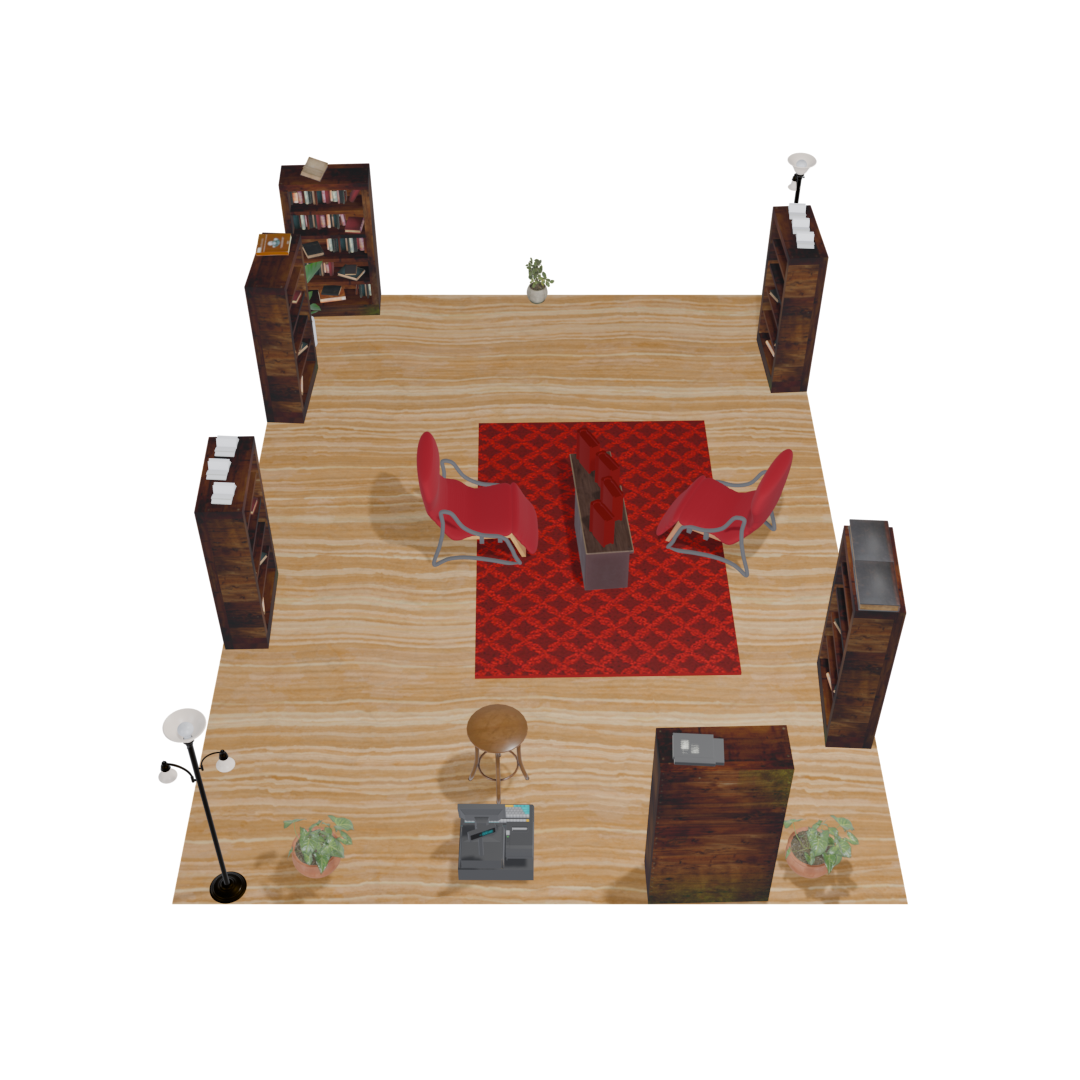}
        \vspace{-28pt}
        \caption{\method}
        \label{fig:method}
    \end{subfigure}
    \hfill
    \begin{subfigure}{0.45\linewidth}
        \centering
        \includegraphics[width=\textwidth]{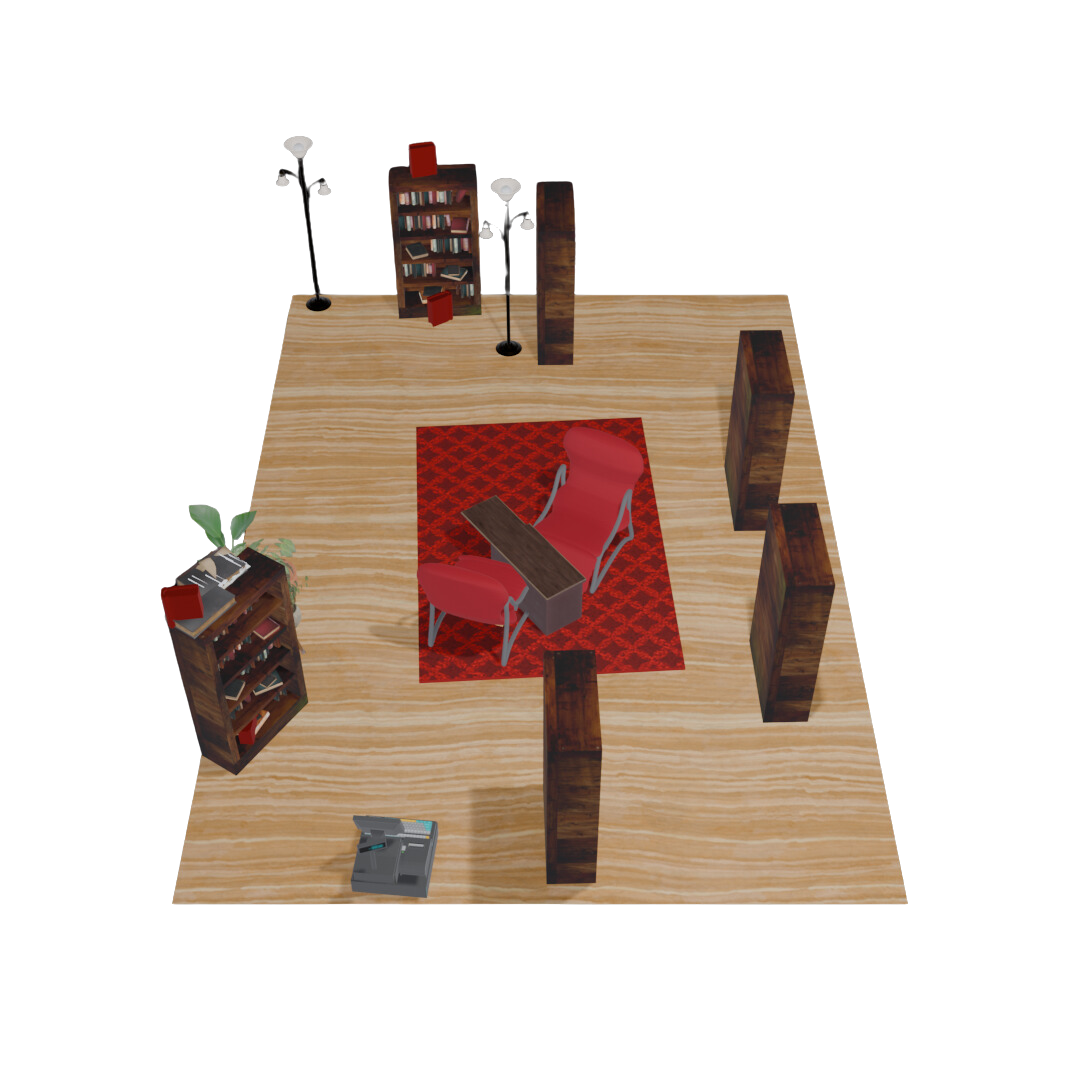}
        \vspace{-28pt}
        \caption{w/o self-consistency decoding}
        \label{fig:method_wo_selfconsistency}
    \end{subfigure}
    \vspace{-8pt}
    \caption{Comparison when w/ and w/o self-consistency decoding.}
    \label{fig:sc_comparison}
\end{figure}

In \Cref{fig:qualitative}, we present qualitative examples of layouts generated by each method. Our method finds valid placements for large numbers of assets. For example, in the florist shop, it places over 10 plants on the central table as instructed. In the buffet restaurant, only our method separates tables, places chairs around each table, and arranges plates on top. Examining the baselines, LayoutGPT produces meaningful placements but often results in object collisions (e.g., a piano colliding with a chair in the dining room) and objects placed outside the room boundary (e.g., bookshelves in the classroom). Holodeck struggles to find reasonable placements for large asset counts. The numerical difference in PSA scores between our method and the best-performing baseline, I-Design, is evident in the qualitative examples.

We further illustrate our model's ability to generate different layouts in response to user instructions, as shown in \Cref{fig:multi_layout}. Our approach demonstrates the ability to closely follow detailed language instructions to generate different layouts even for the same set of assets. The examples show that our method can meet specific user requirements, such as placing chairs on opposite sides or aligning tables in a row. Additionally, our method can generate less conventional layouts, such as placing chairs on top of tables or stacking tables, showcasing its flexibility in interpreting and executing unique instructions.

\subsection{Human Eval to validate our metrics} We follow our baseline I-Design, evaluating layouts using physical plausibility metrics and GPT-4o-rated scores. To further evaluate the alignment between GPT-4o ratings and human ratings, we conducted a user study. %, as well as to provide a human evaluation of the methods. 
Following prior work~\cite{wu2024gpt}, we recruited five graduate students to rank the methods based on position, orientation, and overall performance. They were given the same instructions used as prompts for the GPT-4o evaluator. We collected 495 ratings for each method and metric pair. We converted GPT-4o scores to rankings and computed Kendall’s Tau~\cite{kendall1938new} %to measure alignment both among users and between users and GPT-4o. 
in Table~\ref{tab:alignment}, showing \textbf{strong user-user agreement and user-GPT agreement}. Table~\ref{tab:user_study} reports the average rankings based on user and GPT-4o ratings, demonstrating \textbf{strong correlation with the results in our proposed metrics}.
% % over 11 room types * 3 instances * 3 trials * 4 users * 3 metrics * 4 methods,

%\begin{table}[t]
%    \centering\small
%    % \setlength{\tabcolsep}{1pt}
%    \caption{\textbf{Ablation.} xxx, highlighted all the th}
%    \label{tab:pretrained_ablation}
%    \resizebox{\columnwidth}{!}{%
%    \begin{tabular}{lccccc}
%    \toprule
%     & \multicolumn{2}{c}{Physics} & \multicolumn{2}{c}{Semantics} & \multicolumn{1}{c}{Overall Score} \\
%    \cmidrule(lr){2-3} \cmidrule(lr){4-5}  \cmidrule(lr){6-6}
%    & Collision-free. & Boundary. & Pos. & Rot. & Alignment\\
%    \midrule
%    \model & 81.8 $\pm$ 4.3 &  94.9 $\pm$ 3.8 &  77.5 $\pm$   1.6 &  73.1 $\pm$   1.9 &  58.8 $\pm$ 5.8\\
%    \midrule
%   w/o Visual &  77.8 $\pm$ 5.0 & 87.4 $\pm$   5.0 &  80.2 $\pm$ 2.1 &  75.5 $\pm$ 1.2 & 49.0 $\pm$   5.9\\
%    w/o Self-Consistency & 71.2 $\pm$ 4.3 &  92.9 $\pm$ 3.8 &  75.5 $\pm$ 1.8 &  72.1 $\pm$   1.2 &  46.4 $\pm$   4.1   \\
%    w/o Const. Prog. &  75.8 $\pm$   4.3 &  14.1 $\pm$   5.2 &  70.4 $\pm$   2.3 &  67.3 $\pm$   1.5 &   6.7 $\pm$   2.7  \\
%    \bottomrule
%    \end{tabular}}
%\end{table}

\begin{table}[t]
    \centering%\small
    \resizebox{\columnwidth}{!}{%
    %\scriptsize
    %\setlength{\tabcolsep}{1.5pt}
    \begin{tabular}{lccccc}
    \toprule
     & \multicolumn{2}{c}{Physics} & \multicolumn{2}{c}{Semantics} & \multicolumn{1}{c}{Overall Score} \\
    \cmidrule(lr){2-3} \cmidrule(lr){4-5}  \cmidrule(lr){6-6}
    & CF & IB & Pos. & Rot. & PSA\\
    \midrule
    \method & \textbf{81.8 $\pm$ 2.5} &  \textbf{94.9 $\pm$ 2.2} &  \textbf{77.5 $\pm$   0.9} &  \textbf{73.1 $\pm$ 1.1} &  \textbf{58.8 $\pm$ 3.4}\\
  \midrule
         \multicolumn{6}{l}{\textit{Ablating the key components in LayoutVLM}} \\
         \midrule
   w/o Visual Image &   67.2 $\pm$   1.1 &  \textbf{94.9 $\pm$   0.8} &  \textbf{79.1 $\pm$   1.3} &  \textbf{74.0 $\pm$ 1.2} &  48.2 $\pm$ 1.5 \\
    w/o Self-Consistency & 71.2 $\pm$   2.5 &  \textbf{92.9 $\pm$ 2.2} &  75.5 $\pm$   1.1 &  72.1 $\pm$   0.7 &  46.4 $\pm$   2.4  \\
    w/o Const. & 75.8 $\pm$   2.5 &  14.1 $\pm$   3.0 &  70.4 $\pm$   1.3 &  67.3 $\pm$   0.9 &   6.7 $\pm$   1.6 \\
    \midrule
         \multicolumn{6}{l}{\textit{Ablating the two types of visual marks}} \\
         \midrule
         w/o Visual Asset Mark & 79.8 $\pm$   0.6 &  87.9 $\pm$   3.9 &  \textbf{77.7 $\pm$   0.6} &  72.2 $\pm$   0.8 &  52.2 $\pm$   5.0   \\
    w/o Visual Coordinate  &  77.2 $\pm$   2.0 &  88.9 $\pm$   1.7 &  74.3 $\pm$   1.0 &  68.9 $\pm$   1.3 &  46.0 $\pm$   2.7 \\
    w/o Any Visual Mark &  74.2 $\pm$   6.4 &  84.9 $\pm$   3.1 &  74.5 $\pm$   0.8 &  69.2 $\pm$   0.3 &  43.0 $\pm$   2.2 \\
     \midrule
     \multicolumn{6}{l}{\textit{Ablating the proposed scene layout representation}} \\
     \midrule
    w/o Numerical Init. & 74.7 $\pm$   4.0 &  86.6 $\pm$   2.6 &  69.5 $\pm$   0.1 &  65.4 $\pm$   1.4 &  41.0 $\pm$   1.7 \\
    w/o (Spatial) Const.  & 75.8 $\pm$   2.5 &  14.1 $\pm$   3.0 &  70.4 $\pm$   1.3 &  67.3 $\pm$   0.9 &   6.7 $\pm$   1.6 \\
    %\hline
    \bottomrule
    \end{tabular}
    }
    \caption{\textbf{Ablation.} Removing visual input impedes spatial planning, while omitting self-consistency and optimization negatively impacts both physical feasibility and layout coherence. We also ablate the importance of both types of visual marks and the two key components in our proposed scene layout representation.}
    \label{tab:pretrained_ablation}
    
\end{table}

\begin{table}[t]
    \centering%\small
    %\scriptsize
    %\setlength{\tabcolsep}{1pt}
    \resizebox{\columnwidth}{!}{%
        \begin{tabular}{lccccc}
        \toprule
         %& \multicolumn{5}{c}{} \\ %& \multicolumn{2}{In Distribution}
        %\cmidrule(lr){2-3} \cmidrule(lr){4-5}  \cmidrule(lr){6-6}
       % \multirow{2}{*}{\shortstack{Scene Layout Benchmark \\ (Residential)}} & 
       & \multicolumn{2}{c}{Physics}  & \multicolumn{2}{c}{Semantics} & \multicolumn{1}{c}{Overall}\\
       \cmidrule(lr){2-3} \cmidrule(lr){4-5} \cmidrule(lr){6-6} %\cmidrule(lr){8-8}
        & CF & IB. & Pos. & Rot. & PSA \\
        \midrule
        GPT-4o  &  66.7 $\pm$   5.2 &  \textbf{92.6 $\pm$ 3.0} &  \textbf{71.4 $\pm$ 3.1} &  62.1 $\pm$   2.4 &  \textbf{43.2 $\pm$ 7.0}\\
        GPT-4o (FT on numerical) & \textbf{77.8 $\pm$ 5.2} &  29.6 $\pm$   3.0 &  \textbf{73.8 $\pm$   4.2} &  \textbf{68.5 $\pm$  1.9} &  11.9 $\pm$   0.9 \\
        GPT-4o (FT on ours)  & \textbf{77.8 $\pm$   3.0} &  \textbf{96.3 $\pm$   1.7} &  68.2 $\pm$   0.9 &  62.5 $\pm$ 0.8 &  \textbf{48.1 $\pm$ 1.8}\\
        \midrule
        LLaVA (random) &  66.7 $\pm$   0.0 &   3.7 $\pm$   3.0 &  55.6 $\pm$   1.7 &  47.5 $\pm$   3.0 &   0.7 $\pm$   0.6 \\
        LLaVA (FT on numerical)  & 77.8 $\pm$   5.2 &  18.5 $\pm$   3.0 &  68.4 $\pm$   0.8 &  \textbf{66.0 $\pm$   1.3} &   6.8 $\pm$   2.2\\
        LLaVA (FT on ours) &  \textbf{85.2 $\pm$   6.0} &  \textbf{70.4 $\pm$   8.0} &  \textbf{73.8 $\pm$   1.0} &  \textbf{66.4 $\pm$   0.5} &  \textbf{39.5 $\pm$ 5.7} \\ 
        \bottomrule
        \end{tabular}}
        \caption{\textbf{Comparison of Fine-Tuning VLMs on Different Layout Representations}. FT denotes fine-tuning. Fine-tuning with our representation yields better results than using numerical poses, with significant improvements in the open-source model's performance. }
        \label{tab:finetuning_ablation}
    %}
\end{table}

\subsection{Ablation Study}
%\sun{talk about how with image input, the VLM finds better initialization such that the }

We conduct an ablation study to assess the key components of our approach. In w/o Visual, the VLM was replaced with an LLM. The ablation w/o Self-Consistency removes the step of validating predicted constraints with raw poses, while w/o Const. placed objects based solely on predicted poses, without performing optimization. As shown in \Cref{tab:pretrained_ablation}, the results confirm the effectiveness of our method’s design. Although w/o Visual still uses the same optimization process, which improves physical feasibility, we observe a clear advantage in using a VLM. This likely stems from the VLM’s ability to leverage rendered scenes and asset views to enhance spatial planning (e.g., arranging groups of objects in distinct regions). The comparison with w/o Const. confirms that VLMs alone were insufficient; our scene representation—consisting of both numerical object poses and spatial relations-is essential for generating practical layouts. Finally, we observe that self-consistency takes advantage of our scene representation to help ensure that placements met both physical and semantic requirements. We also analyze the effects of two key components in our proposed scene layout representation. The results indicate that \textbf{spatial constraints are crucial for ensuring physical validity}, particularly in preventing object overlap, while a good numerical initialization contributes to better semantic scores.% Both ablations use code as the underlying representation, demonstrating that the improvements in our method are \textbf{not solely due to replacing CSS or scene graphs with program code}.

\subsection{Finetuning LayoutVLM}
In this experiment, we investigate whether fine-tuning pre-trained VLMs with our scene layout representation improves physical feasibility and semantic coherence. Specifically, we compare this approach with fine-tuning VLMs to directly predict numerical poses. Since the 3D-Front training data consist of typical household room types, we evaluated the fine-tuned models on test cases within the residential category (i.e., bedroom, living room, and dining room). As these test cases include unseen 3D assets from Objaverse and even new object categories, this experiment also assessed generalization capability.
%\subsection{Finetuning Details}
We extract the ground-truth layouts from 3D scenes by selecting defined spatial relations higher than predetermined thresholds. The language instructions are generated based on the type of rooms annotated in the 3D scene dataset. We finetune the open-sourced LLaVA-NeXT-Interleave~\cite{li2024llava} model via LoRA and GPT4-o via the publicly available finetuning API.

The results in \Cref{tab:finetuning_ablation} show that: 1) fine-tuning with our scene representation is more effective than using direct numerical values, and 2) our approach significantly improves the performance of the open-source model. The relatively limited improvement in the closed-source model likely stems from its existing ability to leverage our scene representation through prompt instructions, as shown in previous experiments. In contrast, the open-source VLM struggles with zero-shot layout generation. Fine-tuning with our scene representation enables it to generate better layouts for residential room types with unseen objects.

\section{Conclusion}
%We presented a framework for open-universe 3D layout generation with a. By using VLMs to generate layouts using this coded representation, we show significant improvements over existing methods in terms of 3D layout generation. A few limitations include: \method does not always generate a physically valid layout due to imperfect VLM initializations leading to infeasible convergence. We hope our approach encourages future work to consider more complicated scenes and geometries.
% Similar to existing methods, \method can produce 
%\sun{talk about conclusion, our method still doesn't achieve a 100\% physical plausibility because the VLMs still dont' give perfect initialization, so sometimes the layout gets stuck in local minima (even though it is better than baseline already), cannot consider complex geometries}
In this paper, we present \method, a comprehensive framework for open-universe 3D layout generation, including a novel scene layout representation that builds on two representations of layout from visually marked images: numerical pose estimates and spatial relations. With the scene layout representation combined with self-consistent decoding, differentiable optimization, and visual prompting, we demonstrated significant improvements over existing LLM and constraint-based approaches. While \method shows promise, it has limitations, including occasional failures in generating valid layouts due to suboptimal VLM initializations. We hope our work paves the way for future research to explore more complex scenes and address these challenges, pushing the boundaries of 3D layout generation with improved semantic and physical reasoning.

\paragraph{Acknowledgments.} This work is in part supported by Google, Analog Devices, the Stanford Center for Integrated Facility Engineering (CIFE), the Stanford Institute for Human-Centered AI (HAI), NSF CCRI \#2120095, AFOSR YIP FA9550-23-1-0127, ONR N00014-23-1-2355, and ONR YIP N00014-24-1-2117.

\newpage
{
    \small
    \bibliographystyle{unsrt}
    \bibliography{main}
}

\clearpage
\setcounter{page}{1}
\maketitlesupplementary
\appendix

%\input{table-text/grammar}
%\section{Rationale}
%\label{sec:rationale}
% 
% Having the supplementary compiled together with the main paper means that:
% % 
% \begin{itemize}
% \item The supplementary can back-reference sections of the main paper, for example, we can refer to \cref{sec:intro};
% \item The main paper can forward reference sub-sections within the supplementary explicitly (e.g. referring to a particular experiment); 
% \item When submitted to arXiv, the supplementary will already included at the end of the paper.
% \end{itemize}
% % 
% To split the supplementary pages from the main paper, you can use \href{https://support.apple.com/en-ca/guide/preview/prvw11793/mac#:~:text=Delete%20a%20page%20from%20a,or%20choose%20Edit%20%3E%20Delete).}{Preview (on macOS)}, \href{https://www.adobe.com/acrobat/how-to/delete-pages-from-pdf.html#:~:text=Choose%20%E2%80%9CTools%E2%80%9D%20%3E%20%E2%80%9COrganize,or%20pages%20from%20the%20file.}{Adobe Acrobat} (on all OSs), as well as \href{https://superuser.com/questions/517986/is-it-possible-to-delete-some-pages-of-a-pdf-document}{command line tools}.

\newpage
\section{Details of \method}
This section elaborates on the details of our method, including the prompts we used.

\subsection{Grouping}
To address the challenge of handling large numbers of 3D assets, we cluster related assets into groups using the following prompt:
\styledfileinput[python]{program/grouping.txt}{}
%\styledfileinput{program/prompt.txt}
%\subsection{Semantic Asset Group}
%\subsection{Visual Mark}

\subsection{Differentiable Spatial constraint}
%\section{Differentiable Objectives}
\label{supp:loss}
We define differentiable objectives for the spatial constraints used in our method. We introduce the necessary notations and provide the mathematical formulations below. The pose of an asset \( m_i \) is represented as \( p_i = (x_i, y_i, z_i, \theta_i) \), the orientation of the asset $\theta_i$ is represented with an orientation vector  \( \mathbf{v}_i = (\cos\theta_i, \sin\theta_i) \), and \( b_i \) denotes the 3D bounding box size of the asset. Below are the mathematical objectives for various spatial relations.
\paragraph{Distance Objective}

\begin{align}
&\mathcal{L}_{\text{distance}}(p_i, p_j, d_{\text{min}},  d_{\text{max}}) = \text{clamp}\big(\\
&\min(\|p_i - p_j\| - d_{\text{min}}, \notag d_{\text{max}} - \|pos_i - pos_j\|), 0, 1 \big)
\end{align}

where \( \|pos_i - p_j\| = \sqrt{(x_i - x_j)^2 + (y_i - y_j)^2} \) is the Euclidean distance between \( i \) and \( j \) in the x-y plane. The function \(\text{clamp}(x, a, b)\) constrains \( x \) to \([a, b]\), defined as \(\text{clamp}(x, a, b) = \min(\max(x, a), b)\).

\paragraph{On-Top-Of Objective}

\begin{equation}
\mathcal{L}_{\text{on\_top\_of}}(p_i, p_j, b_i, b_j) = -\mathcal{L}_{\text{DIoU}}(p_i, p_j, b_i, b_j).
\end{equation}
where \(\text{IoU}(a, b)\) denotes the Intersection-over-Union of the bounding boxes of \( a \) and \( b \). Instead of using a loss function for the z-axis, the On-Top-Of objective directly sets  \( z_i \) the z-coordinate of the object $i$ to be on top of the object \( j \).

\paragraph{Point-Towards Objective}

\begin{align}
\mathcal{L}_{\text{point\_towards}}(p_i, p_j, \phi) = \begin{cases}
    0, & \text{if } \mathbf{v}_i \cdot \mathbf{d}_{ij} > 0, \\
    1 - \frac{\mathbf{v}_i \cdot \mathbf{d}_{ij}}{\|\mathbf{v}_i\| \|\mathbf{d}_{ij}\|}, & \text{otherwise,}
\end{cases}
\end{align}

where \( \mathbf{d}_{ij} \) is the direction vector from \( i \) to \( j \) rotated by $\phi$ degree around the $z$-axis.
%, and \( \mathbf{v}_i = (\cos\theta_i, \sin\theta_i) \) is the orientation vector of \( i \).

\paragraph{Align-With Objective}

\begin{equation}
\mathcal{L}_{\text{align\_with}}(p_i, p_j, \phi) = 1 - \frac{\mathbf{v}_i \cdot \mathbf{v}_j}{\|\mathbf{v}_i\| \|\mathbf{v}_j\|},
\end{equation}

where \( \mathbf{v}_i \) and \( \mathbf{v}_j \) are the orientation vectors of \( i \) and \( j \), respectively.

\paragraph{Against-Wall Objective}

The "Against-Wall" objective consists of two components: (a) the sum of distances from the object's corners to the wall, and (b) a term that encourages the object to point away from the wall. Let \( c_i^{(k)} \) be the \( k \)-th corner of the \( i \)-th object, which is calculated based on the object's $x$-$y$ position \((x_i, y_i)\), rotation, and bounding box size \(b_i\). The loss function is:

\begin{align}
\mathcal{L}_{\text{against\_wall}}(p_i, w_j, b_i) = & \sum_{k=1}^{4} \text{clamp}\big(\|c_i^{(k)} - w_j\|, 0, 1\big) \notag \\
& + \left( 1 - \frac{\mathbf{v}_i \cdot \mathbf{n}_{w_j}}{\|\mathbf{v}_i\| \|\mathbf{n}_{w_j}\|} \right),
\end{align}

where $\|c_i^{(k)} - w_j\|$ denotes the Euclidean distance between the $k$-th corner of  to the wall segment on the $x$-$y$ plane and \( \mathbf{n}_w \) denotes the normal vector of wall \( w \) (i.e., perpendicular to the wall).

\paragraph{Optimization Details}
%\underline{\textbf{Optimization Details}}\newline
%\underline{\textbf{(\Rtwo, \Rthree, \Rfour).}}\newline
We use Adam optimizer and Exponential LR scheduler with a decay factor $0.96$. Each optimization runs for 400 steps with projection back to the boundary every 100 iterations. Optimizing a scene with ~40 assets takes 1–5 minutes on a single GPU, 5-10 GPT-4o calls (i.e., one per group), varying based on the VLM-defined optimization problem.

Below is the prompt we feed VLM to generate spatial constraints. 
\styledfileinput[python]{program/prompt.txt}{}

\subsection{Self-Consistent Decoding}
We propose self-consistent decoding to address the challenge of maintaining layout coherence in VLM-generated spatial plans. Our main hypothesis is that preserving self-consistent spatial relations—those that align with the estimated numerical poses of objects—is essential for ensuring semantic and physical plausibility during optimization. During implementation, we simplify the decoding process by enforcing that each asset maintains at most one orientational constraint, either to ``point towards" or ``align with" another asset. Additionally, the spatial relation "on top of" is excluded from the self-consistency decoding, as we empirically observe that ``on top of " relations are almost accurately and reasonably predicted by our model; thus, enforcing self-consistency is unnecessary.

\subsection{Annotating Unlabeled 3D Assets}
%\subsection{GPT-4-V for 3D Asset Annotation}
We annotate the 3D assets used in a similar way as in Holodeck~\cite{yang2024holodeck}, using GPT-4o
to determine the front face of the object and to determine the textual description of the asset. More specifically, GPT-4o takes a set of four images as inputs, each showing an object from orthogonal rotations (0°, 90°, 180°, and 270°) and outputs the following attributes for the 3D object:
\begin{itemize}
\item \textbf{Category}: a specific classification of the object, such as ``chair'', ``table'', ``building'', etc.
\item \textbf{Variable Name}: a string denoting the python variable name that will be used to refer to this object in our scene layout representation.
\item \textbf{Front View}: an integer denoting the view representing the front of the object, often the most symmetrical view.
\item \textbf{Description}: a detailed textual description of the object.
\item \textbf{Materials}: a list of materials constituting the object.
\item \textbf{Placement Attributes:} Boolean values (\textsc{onCeiling}, \textsc{onWall}, \textsc{onFloor}, \textsc{onObject}) indicating typical placement locations. For example, ``True'' for a ceiling fan's placement on the ceiling.
\end{itemize}

\section{Details of our Experiments}
\subsection{Generating Test Cases}
We developed a pipeline for generating valid open-vocabulary 3D layout generation cases to benchmark our method against existing methods.%, for retrieving Objaverse~\cite{deitke2023objaverse} assets given room types.
 
First, we feed the following prompt to GPT-4o to generate a layout instruction given the room type:
\styledfileinput[python]{program/task_generate_criterion.txt}{}

Condition on the generated room layout instruction, we then use the following prompt to retrieve a bunch of plausible assets:
\styledfileinput[python]{program/task_generate_asset.txt}{}

Subsequently, we embed the generated asset descriptions using CLIP~\cite{radford2021learning} and use the embeddings to retrieve 3D assets from Objaverse. The following prompt is employed to verify whether the retrieved object belongs to the given room:
\styledfileinput[python]{program/task_generate_verifier.txt}{}
At last, we conduct many verifications to remove assets that humans deem unsuitable given the room type and layout instruction (e.g., a 3D asset of an entire city should not appear in an indoor scene).

\subsection{Evaluation}
Evaluating the quality of generated 3D layouts requires metrics that measure both physical plausibility and semantic coherence. In this section, we introduce the evaluation prompts we feed to VLM to assess the performance of layout generation systems.  

We measure the positional and rotational \textit{Semantic coherency} score with the following prompts. 
\styledfileinput[python]{program/position.txt}{}

\styledfileinput[python]{program/rotation.txt}{}

We measure the \textit{Physically-grounded Semantic Alignment Score (PSA)} with \textit{Collision-Free Score (CF)}, \textit{In-Boundary Score (IB)}, and the overall prompt alignment score following prompt. 
\styledfileinput[python]{program/metrics.txt}{}

\section{More Qualitative Comparison}

\begin{figure*}
    \centering
    \includegraphics[width=\textwidth]{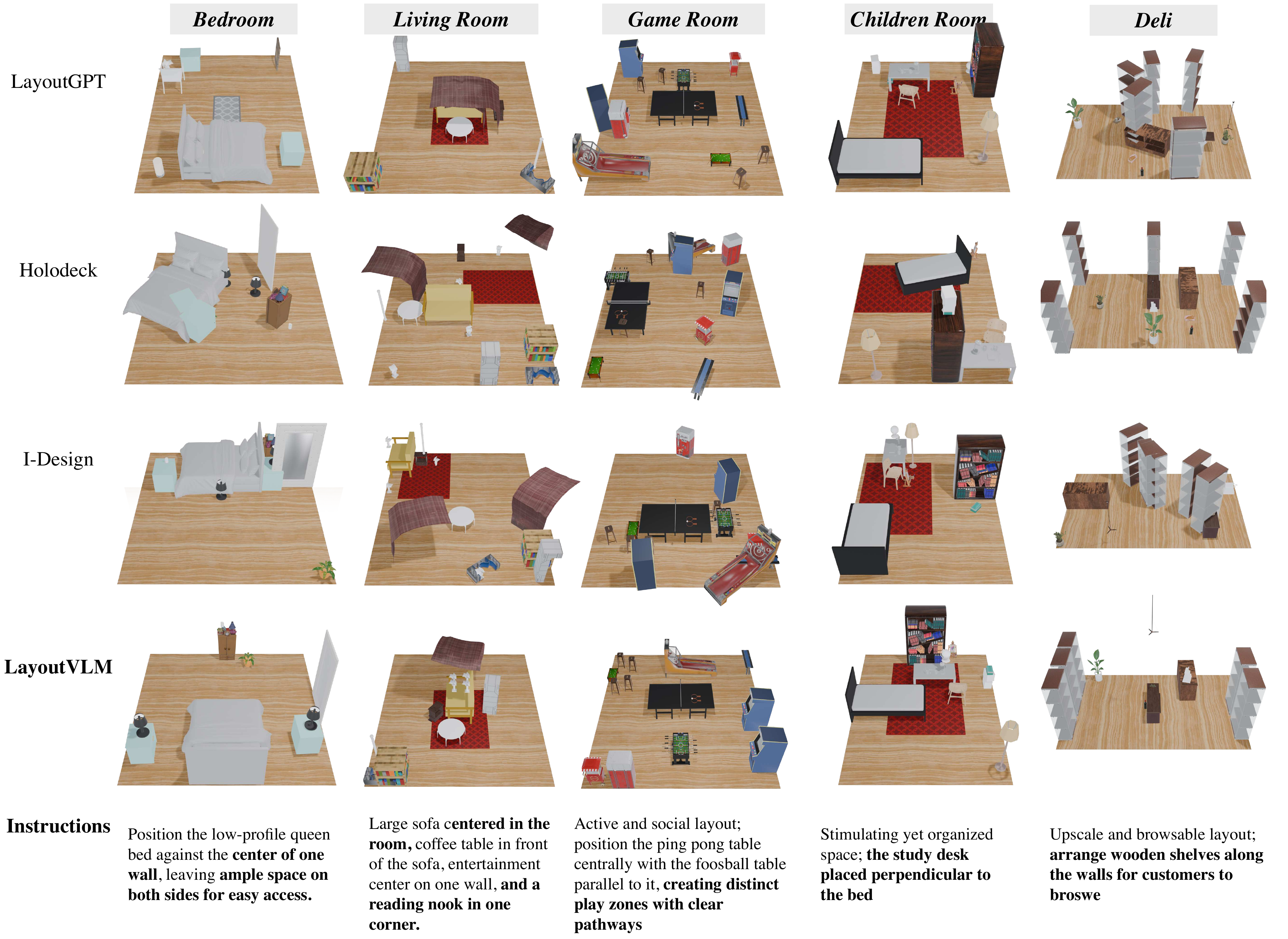}
    %\vspace{-0.5cm}
    \caption{More qualitative comparison with baseline methods in generating layouts based on detailed language instructions. }
    \label{fig:qualitative2}
    %\vspace{-0.2cm}
\end{figure*}

In \Cref{fig:qualitative2}, we present more qualitative examples of layouts generated by \method and baseline methods. \method consistently outperforms baseline methods across all room types in terms of both physcial plausibility and semantic coherency. In the Living Room example, \method excels in identifying semantic asset group by clustering sofa and table together. By leveraging spatial reasoning from VLMs, we are able to stack assets toegther, as shown in the Deli example.

\end{document}